\documentclass{article}

\PassOptionsToPackage{numbers, compress}{natbib}
\usepackage[preprint]{neurips_2026}

\usepackage[utf8]{inputenc} 
\usepackage[T1]{fontenc}    
\usepackage{hyperref}       
\usepackage{url}            
\usepackage{booktabs}       
\usepackage{amsfonts}       
\usepackage{nicefrac}       
\usepackage{microtype}      
\usepackage{xcolor}         

\usepackage{graphicx}
\usepackage[table]{xcolor}
\usepackage{caption}
\newcommand{\best}[1]{\textbf{#1}}
\newcommand{\second}[1]{\underline{#1}}
\usepackage{wrapfig}
\usepackage{booktabs}
\usepackage{subcaption}
\usepackage{adjustbox}
\usepackage{url}

\usepackage{xcolor}

\definecolor{dyncolor}{RGB}{230,126,72}   
\definecolor{statcolor}{RGB}{72,156,208}  
\definecolor{n2scolor}{RGB}{96,60,137}
\definecolor{s2ncolor}{RGB}{116,33,104}

\title{\textsc{SpaceNum}: Revisiting Spatial Numerical Understanding in VLMs}

\author{%
  Jianshu Zhang\thanks{Equal contribution. Project page: \url{https://sterzhang.github.io/SpaceNum-Home/}} \\
  Northwestern \\
  \And
  Yijiang Li\footnotemark[1] \\
  UCSD \\
  \And
  Huifeixin Chen \\
  USC \\
  \And
  Haoran Lu \\
  Northwestern  \\
  \AND
  Letian Xue \\
  Northwestern \\
  \And
  Bingyang Wang \\
  GaTech \\
  \And
  Han Liu \\
  Northwestern \\
}

\begin{document}

\maketitle

\begin{abstract}
Vision-Language Models (VLMs) are increasingly deployed in embodied environments, where they need produce numerical outputs such as action magnitudes and spatial coordinates. Although these numbers appear meaningful, it remains unclear whether these numerical outputs are genuinely grounded in spatial perception. 
Therefore, in this work, we revisit spatial numerical understanding through \textsc{\textbf{SpaceNum}}, a unified framework that captures two complementary settings: numbers as dynamic transitions during spatial exploration, and numbers as static layouts in spatial reasoning. We formulate two bidirectional tasks, \textsc{Num2Space} and \textsc{Space2Num}, to evaluate how well VLMs map between vision-side spatial structure and language-side numerical representations.
We systematically study whether current VLMs truly understand numerical values in spatial settings. Across dynamic transitions and static layouts, we find that models largely fail to ground numbers in spatial meaning and often perform close to random guess. Through error analysis, reasoning trace analysis, and controlled interventions, we show that current VLMs rely heavily on shallow spatial cues, struggle to build stable coordinate-aware representations, and fail to abstract structured spatial layouts from visual observations. We further show that explicit reasoning provides only marginal gains, while tuning can partially improve spatial numerical understanding and transfer to external spatial reasoning benchmarks.
\end{abstract}

\section{Introduction}
\begin{figure}[t]
  \centering
  \includegraphics[width=1\linewidth]{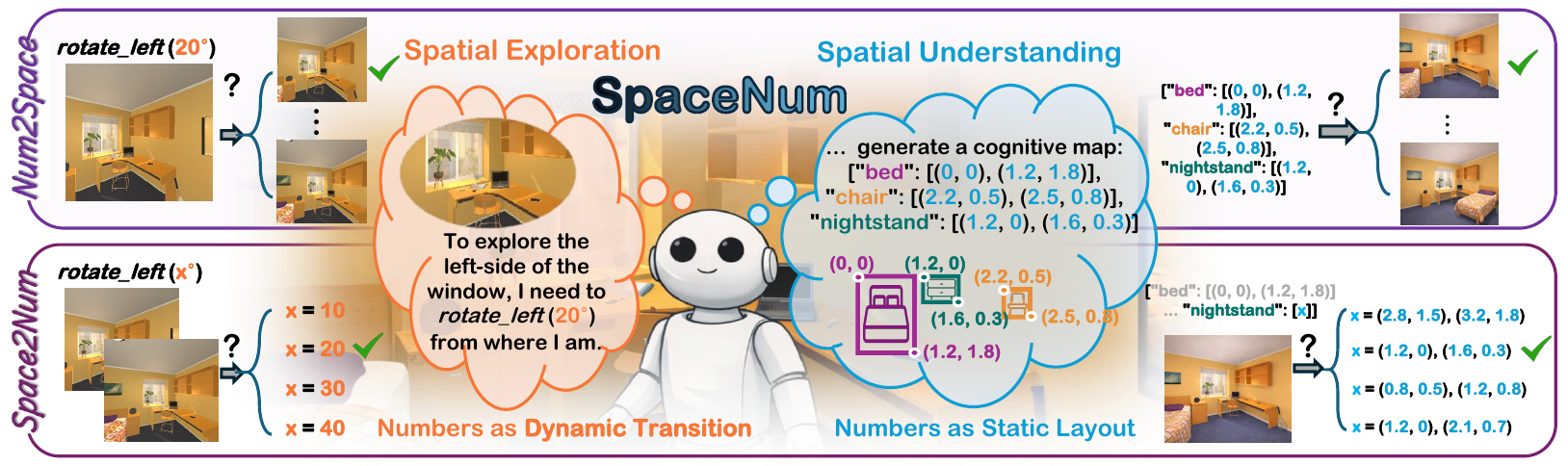}
  \caption{
Overview of \textsc{\textbf{SpaceNum}}. 
We study spatial numerical understanding under two settings: 
\textcolor{dyncolor}{\textit{numbers as dynamic transition}} in spatial exploration (left) and 
\textcolor{statcolor}{\textit{numbers as static layout}} in spatial understanding (right). 
We further investigate the mapping between vision-side space and language-side numbers via two tasks: 
\textcolor{n2scolor}{\textsc{Num2Space}}, which maps numbers to visual outcomes (top), and 
\textcolor{s2ncolor}{\textsc{Space2Num}}, which maps visual inputs to numbers (bottom).
}
  \label{fig:teaser}
\end{figure}

Vision-language models (VLMs) have recently progressed from describing what is directly visible in images~\citep{dai2023instructblip, liu2023visual, pi2024image} to actively exploring and understanding complex spatial environments~\citep{ray2024sat, jia2025omnispatial, yang2025thinking, du2024embspatial, ma2025spatialreasoner}. Two representative spatial task scenarios have emerged: (1) \textbf{spatial exploration}, where a VLM-based agent navigates an environment by generating actions conditioned on its observations to actively gather information; and (2) \textbf{spatial understanding}, where VLMs infer the global structure of a scene and answer spatially grounded questions by constructing an internal representation of the environment. 
As illustrated in Figure~\ref{fig:teaser}, despite their different objectives, both paradigms share a common requirement: VLMs must produce explicit numerical values whose meanings are grounded in spatial context.

In spatial exploration~\citep{yang2025mindjourney, wang2026hydra}, a VLM-based agent may output an action such as ``\textit{rotate\_left}(20$^\circ$)''. The value $20$ does not describe the current observation, nor does it directly specify the next observation. Instead, it specifies the magnitude of a state change, serving as a transition quantity between consecutive observations, where numbers naturally function as \textbf{\textit{dynamic transition magnitudes}}.

In contrast, in spatial understanding, prior work has shown that constructing explicit spatial representations~\citep{yin2025spatial, yang2025thinking, huang2025video2layout}, often in the form of cognitive maps, improves performance on spatial reasoning tasks. Here, numbers encode relative spatial relationships and correspond to \textbf{\textit{static relative spatial layouts}}. A single object's coordinates in isolation carry limited semantic meaning; spatial information becomes interpretable only when multiple objects are considered within a shared coordinate system, where numerical values define their relative positions and overall layout.

This naturally raises a key question: \textbf{do VLMs genuinely understand numbers as metric quantities in space and generate them grounded in metric properties of space?}
Across both spatial exploration and spatial understanding, \textsc{\textbf{Num2Space}} evaluates whether a language-side numerical value can be correctly grounded in its corresponding spatial outcome, while \textsc{\textbf{Space2Num}} tests whether an appropriate numerical value can be inferred from a given spatial configuration. Together, these two tasks assess numerical understanding from both directions, enabling a systematic examination of whether VLMs merely generate plausible numbers or genuinely ground them in spatial meaning.

To systematically study spatial numerical understanding, we investigate a series of progressively deeper questions. We first evaluate 18 VLMs across dynamic transitions and static layouts, showing that current models largely fail to ground numerical values in spatial meaning and often perform close to random guess. We then analyze how these failures differ across scenarios and mapping directions, revealing strong asymmetries between vision-to-number and number-to-vision grounding. To further understand the source of these failures, we conduct structured error analysis, reasoning trace analysis, and controlled interventions. Our results show that current VLMs often rely on shallow spatial cues, fail to construct stable coordinate-aware representations, and struggle to abstract structured spatial layouts from visual observations. Surprisingly, enabling explicit reasoning brings only marginal improvements, suggesting that the main limitation is not the absence of reasoning traces, but the lack of spatially calibrated reasoning operations. Finally, we show that spatial numerical understanding can be partially improved through tuning and transfers to external spatial reasoning benchmarks.

\section{SpaceNum Data Curation}

\paragraph{Data Source and Platform.}
We setup simulator-based pipelines to enable controllable data generation. For dynamic transition, data is generated in AI2-THOR~\citep{kolve2017ai2}, which supports embodied agents executing parameterized actions across diverse indoor environments. For static layout data, scenes are built in NVIDIA Isaac Sim~\citep{nvidia_isaac_sim} using assets from BlenderKit~\citep{blenderkit}, allowing controlled layout generation with access to ground-truth spatial annotations for cognitive map construction.

\subsection{Number as Dynamic Transition}

\paragraph{Data Collection.}
We construct dataset with careful control over action coverage, transition continuity, visual anchoring, and data validity. 
(i) \textbf{Action coverage:} We define a set of primitive actions that induce spatial transitions, including \texttt{Move Forward (F) / Backward (B); Left (L) / Right (R)}) and rotations (\texttt{Rotate Up (U) / Down (D); Left (L) / Right (R)}. 
(ii) \textbf{Transition continuity:} The action magnitudes are chosen to ensure sufficient overlap between consecutive observations, as summarized in Table~\ref{tab:action_range}, maintaining \textit{visual continuity} while introducing meaningful spatial changes and avoiding abrupt or ambiguous transitions. 
\begin{wraptable}[8]{r}{0.39\linewidth}
\centering
\small
\begin{tabular}{lcc}
\toprule
\textbf{Action} & \textbf{Range} & \textbf{Step} \\
\midrule
\texttt{Move F/B} & 0.2--2.4 m & 0.2 \\
\texttt{Move L/R} & 0.2--1.2 m & 0.2 \\
\texttt{Rotate U/D} & 10--70° & 10 \\
\texttt{Rotate L/R} & 10--70° & 10 \\
\bottomrule
\end{tabular}
\caption{Action parameter ranges.}
\label{tab:action_range}
\end{wraptable}
(iii) \textbf{Visual anchoring:} To ensure transitions are visually identifiable, we filter out observations with insufficient anchors by discarding frames containing fewer than \textit{3 object instances}. 
(iv) \textbf{Data validity:} To avoid invalid transitions caused by random initialization or action execution (e.g., identical frames or empty observations), we leverage \textit{occupancy maps} to constrain both the initial agent state and the post-action state to be valid, ensuring all collected samples correspond to informative transitions.

\paragraph{Task Definition.}
Let $o_t$ denote the initial observation, $o_{t+1}$ the resulting observation, $a$ the action type, and $n$ the numerical parameter representing the transition magnitude.

\textbf{\textsc{Num2Space}.} 
The model is given $(o_t, a, n)$ and is required to select the correct resulting observation $o_{t+1}$ from a set of candidates. The distractor candidates are constructed by fixing the same initial observation $o_t$ and action type $a$, while varying the numerical value $n$, resulting in alternative observations $\tilde{o}_{t+1}$ that correspond to different transition magnitudes.

\textbf{\textsc{Space2Num}.} 
The model is given $(o_t, o_{t+1}, a)$ and is required to infer the numerical value $n$ that explains the transition. This task requires grounding visual differences between $o_t$ and $o_{t+1}$ to the corresponding transition magnitude.

\subsection{Number as Static Layout}

\paragraph{Data Collection.}
We build the layout dataset with controlled generation, covering the reference system, layout construction, scene scale, and representation.
\textbf{(i) Coordinate system construction.}
Each scene uses a clear coordinate system defined by two anchor objects. One anchor sets the origin. The relative position of the two anchors defines a consistent direction. This fixes the coordinate frame (up to scale) and removes ambiguity. The anchors stay fixed across samples in the same scene.
\textbf{(ii) Layout generation.}
Given the coordinate system, we place a third object with different positions and sizes. We enforce simple constraints: objects do not overlap, and distances are within a reasonable range. Under the same reference frame, we create three types of changes: (a) position only, (b) size only, and (c) both position and size. This lets us study each factor in a controlled way.
\textbf{(iii) Scene scale.}
We include both desktop-scale and room-scale scenes. This changes the spatial extent and the distribution of objects, and adds diversity.
\textbf{(iv) Representation variation.}
For each layout, we build multiple coordinate-based representations with different dimensions (1D, 2D, and 3D). These representations describe the same layout in different forms, from simple to more complete ones. This helps us study how models handle spatial information under different representations.

\paragraph{Task Definition.}
Let $\mathcal{M}$ denote a number-based cognitive map, $o$ the layout observation, and $p$ the numerical coordinates of a target object under a given reference frame.

\textbf{\textsc{Num2Space}.} 
The model is given a cognitive map $\mathcal{M}$ and is required to select the observation $o$ that is consistent with the specified layout. Distractor candidates are constructed by varying object positions or sizes while preserving the same reference frame.

\begin{wrapfigure}[7]{r}{0.48\linewidth}
\vspace{-0.6cm}
  \centering
  \includegraphics[width=\linewidth]{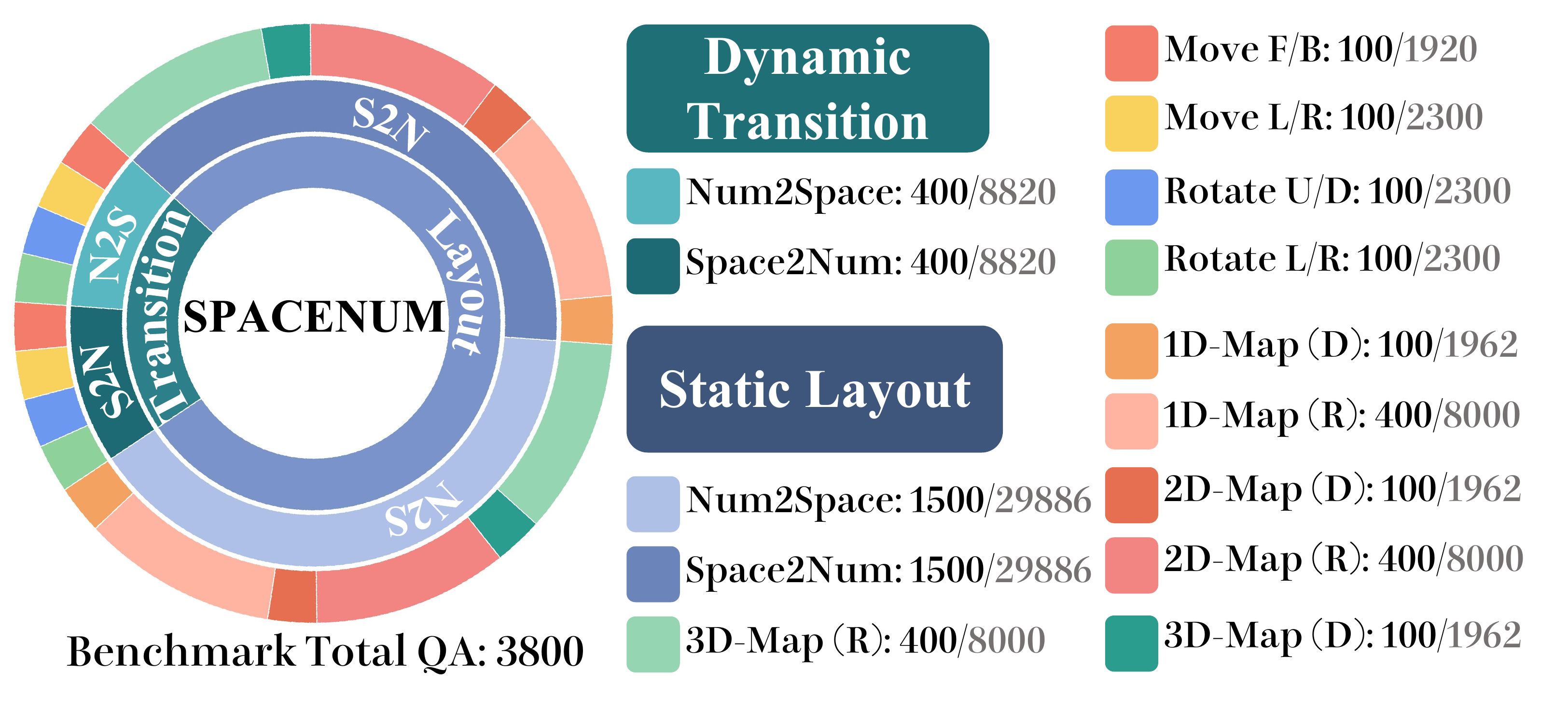}
  \caption{Dataset statistics.}
  \label{fig:statistics}
\vspace{-0.2cm}
\end{wrapfigure}

\textbf{\textsc{Space2Num}.} 
The model is given an observation $o$ and is required to infer the numerical coordinates $p$ of a target object under the reference coordinate system. 
This task requires grounding visual spatial structure into numerical representations.

\subsection{Statistics}

Figure~\ref{fig:statistics} summarizes the benchmark composition that contains 3,800 samples.
We further use the same fully automatic pipeline to generate an additional 77,412 training samples for later training-based explorations. 
The detailed breakdown of this larger training set is also shown in gray in Figure~\ref{fig:statistics}.

\newcommand{\graynum}[1]{\textcolor{gray}{#1}}

\begin{table*}[t]
\centering
\fontsize{4.8pt}{5.3pt}\selectfont
\setlength\tabcolsep{3pt}
\renewcommand{\arraystretch}{1.15}
\begin{adjustbox}{max width=\textwidth}
\begin{tabular}{r|c|c|cccc|cccc|cccccc|cccccc}
& & &
\multicolumn{8}{c|}{\cellcolor{dyncolor!50}\textsc{\textbf{Dynamic Transition}}} &
\multicolumn{12}{c}{\cellcolor{statcolor!50}\textsc{\textbf{Static Layout}}} \\

& & &
\multicolumn{4}{c|}{\textsc{\textbf{Num2Space}}} &
\multicolumn{4}{c|}{\textsc{\textbf{Space2Num}}} &
\multicolumn{6}{c|}{\textsc{\textbf{Num2Space}}} &
\multicolumn{6}{c}{\textsc{\textbf{Space2Num}}} \\

& & &
\multicolumn{2}{c}{\textbf{Move}} & \multicolumn{2}{c|}{\textbf{Rotate}} &
\multicolumn{2}{c}{\textbf{Move}} & \multicolumn{2}{c|}{\textbf{Rotate}} &
\multicolumn{2}{c}{\textbf{1D-Map}} & \multicolumn{2}{c}{\textbf{2D-Map}} & \multicolumn{2}{c|}{\textbf{3D-Map}} &
\multicolumn{2}{c}{\textbf{1D-Map}} & \multicolumn{2}{c}{\textbf{2D-Map}} & \multicolumn{2}{c}{\textbf{3D-Map}} \\

Methods & Rank & Avg. &
\texttt{F/B} & \texttt{L/R} &
\texttt{U/D} & \texttt{L/R} &
\texttt{F/B} & \texttt{L/R} &
\texttt{U/D} & \texttt{L/R} &
\texttt{D} & \texttt{R} & \texttt{D} & \texttt{R} & \texttt{D} & \texttt{R} &
\texttt{D} & \texttt{R} & \texttt{D} & \texttt{R} & \texttt{D} & \texttt{R} \\
\hline

\rowcolor{gray!10}
Random Guess &  & 30.0
& 25.0 & 25.0 & 25.0 & 25.0
& 25.0 & 25.0 & 25.0 & 25.0
& 50.0 & 50.0 & 25.0 & 25.0 & 25.0 & 25.0
& 50.0 & 50.0 & 25.0 & 25.0 & 25.0 & 25.0 \\

\cellcolor{orange!6}Qwen2.5-VL-72B & 1 & 39.8
& 34.0 & \best{38.0} & 34.0 & \best{37.0}
& 40.0 & 37.0 & 44.0 & 41.0
& 69.0 & 64.5 & 28.0 & \graynum{24.2} & \second{36.0} & 26.8
& \second{60.0} & 51.2 & 33.0 & \best{33.8} & 31.0 & 32.8 \\

\cellcolor{blue!6}InternVL3.5-38B & 2 & 39.5
& \best{38.0} & 27.0 & 30.0 & 29.0
& \second{42.0} & 38.0 & \second{47.0} & \second{42.0}
& 69.0 & 52.8 & \best{31.0} & \graynum{24.2} & 35.0 & \graynum{23.2}
& 53.0 & 54.5 & \best{43.0} & \second{32.5} & \second{40.0} & \best{38.2} \\

\cellcolor{orange!6}Qwen2.5-VL-32B & 3 & 38.5
& 32.0 & 30.0 & \second{36.0} & \graynum{22.0}
& 37.0 & 33.0 & 41.0 & 38.0
& \best{71.0} & \best{67.0} & \graynum{25.0} & \graynum{23.2} & \best{37.0} & 25.2
& \best{63.0} & 55.8 & 38.0 & 28.5 & 34.0 & \second{33.2} \\

\cellcolor{blue!6}InternVL3.5-14B & 4 & 38.2
& 36.0 & 32.0 & \best{37.0} & 27.0
& 40.0 & 35.0 & \best{53.0} & \best{48.0}
& \best{71.0} & \second{66.8} & \graynum{20.0} & \graynum{24.0} & 27.0 & 25.5
& 53.0 & 54.8 & 30.0 & 27.5 & 34.0 & \graynum{23.0} \\

\cellcolor{cyan!6}Qwen3-VL-32B & 5 & 35.9
& 26.0 & 30.0 & \second{36.0} & \graynum{25.0}
& 36.0 & \best{49.0} & 44.0 & 32.0
& 68.0 & 50.2 & \second{30.0} & \graynum{20.8} & 32.0 & \graynum{22.8}
& 58.0 & 57.2 & 28.0 & \graynum{23.0} & 29.0 & \graynum{20.8} \\

\cellcolor{blue!6}InternVL3.5-8B & 6 & 34.8
& 30.0 & 28.0 & 35.0 & 29.0
& \best{45.0} & 30.0 & 38.0 & 28.0
& 64.0 & 64.8 & \graynum{21.0} & \graynum{22.5} & 31.0 & \graynum{22.0}
& 53.0 & 52.8 & 36.0 & \graynum{19.2} & \graynum{25.0} & \graynum{20.8} \\

\cellcolor{green!6}Ovis2.5-9B & 7 & 34.7
& \graynum{22.0} & 32.0 & 31.0 & \graynum{23.0}
& 36.0 & \second{44.0} & 41.0 & 27.0
& \second{70.0} & 66.2 & \graynum{17.0} & \graynum{25.0} & \graynum{21.0} & \graynum{24.8}
& 53.0 & \best{58.5} & \graynum{24.0} & 28.7 & 26.0 & \graynum{24.5} \\

\cellcolor{blue!6}InternVL3.5-4B & 8 & 34.5
& 26.0 & 29.0 & \graynum{25.0} & \graynum{21.0}
& 35.0 & 29.0 & 34.0 & 36.0
& \second{70.0} & 61.0 & \second{30.0} & \graynum{23.2} & 30.0 & \graynum{22.8}
& 56.0 & \second{58.2} & 30.0 & \graynum{18.8} & 38.0 & \graynum{18.0} \\

\cellcolor{cyan!6}Qwen3-VL-8B & 9 & 33.4
& 26.0 & 33.0 & 30.0 & \graynum{25.0}
& 35.0 & 33.0 & 43.0 & 30.0
& \graynum{37.0} & \graynum{43.8} & 26.0 & \best{30.0} & \graynum{24.0} & 26.0
& 57.0 & \graynum{49.5} & \second{39.0} & \graynum{22.0} & 35.0 & \graynum{22.8} \\

\cellcolor{green!6}Ovis2.5-2B & 10 & 33.2
& 26.0 & \graynum{22.0} & 29.0 & 31.0
& 27.0 & 27.0 & \graynum{23.0} & \graynum{24.0}
& \best{71.0} & \best{67.0} & 28.0 & \graynum{22.0} & 27.0 & \graynum{24.5}
& 51.0 & \graynum{49.5} & 28.0 & 26.2 & 33.0 & 27.8 \\

\cellcolor{teal!6}Cosmos-Reason2-8B & 11 & 33.1
& \graynum{24.0} & \second{37.0} & 29.0 & \graynum{25.0}
& 31.0 & 26.0 & 27.0 & 33.0
& 57.0 & 53.5 & \graynum{20.0} & \second{28.0} & \graynum{20.0} & 27.0
& 58.0 & 50.7 & 34.0 & \graynum{23.8} & 30.0 & 27.3 \\

\cellcolor{orange!6}Qwen2.5-VL-7B & 12 & 33.0
& \second{37.0} & \graynum{22.0} & 30.0 & \second{32.0}
& 29.0 & 29.0 & 27.0 & 30.0
& \best{71.0} & \best{67.0} & \graynum{21.0} & 26.0 & \graynum{25.0} & \second{27.5}
& \graynum{46.0} & \graynum{47.5} & 29.0 & \graynum{23.5} & \graynum{20.0} & \graynum{20.5} \\

\cellcolor{cyan!6}Qwen3-VL-4B & 13 & 32.1
& \graynum{22.0} & 29.0 & 26.0 & 26.0
& 31.0 & 35.0 & 29.0 & 32.0
& \graynum{41.0} & 55.2 & 28.0 & \graynum{23.5} & \graynum{23.0} & \graynum{24.5}
& 57.0 & 56.0 & 33.0 & \graynum{20.2} & 31.0 & \graynum{19.5} \\

\cellcolor{orange!6}Qwen2.5-VL-3B & 14 & 31.9
& \graynum{24.0} & \graynum{20.0} & \graynum{23.0} & 29.0
& 26.0 & \graynum{16.0} & \graynum{25.0} & \graynum{20.0}
& \best{71.0} & \best{67.0} & \graynum{19.0} & \graynum{24.8} & 30.0 & \best{28.0}
& 55.0 & \graynum{41.5} & 34.0 & 25.5 & \best{41.0} & \graynum{17.8} \\

\cellcolor{teal!6}Cosmos-Reason2-2B & 15 & 31.6
& 28.0 & \graynum{22.0} & \graynum{23.0} & \graynum{25.0}
& \graynum{23.0} & 26.0 & \graynum{24.0} & 26.0
& \best{71.0} & \best{67.0} & \graynum{13.0} & 27.0 & \graynum{13.0} & \graynum{23.5}
& \graynum{48.0} & 55.2 & \graynum{25.0} & 27.0 & 39.0 & 27.3 \\

\cellcolor{purple!6}Gemma-3-27B & 16 & 31.2
& 27.0 & \graynum{25.0} & 34.0 & \graynum{16.0}
& \graynum{24.0} & 29.0 & \graynum{25.0} & 27.0
& \graynum{50.0} & \graynum{43.2} & \graynum{25.0} & \graynum{23.8} & \graynum{22.0} & \graynum{22.8}
& 54.0 & \graynum{49.0} & 32.0 & \graynum{24.5} & \best{41.0} & 29.0 \\

\cellcolor{purple!6}Gemma-3-12B & 17 & 30.6
& \graynum{21.0} & 26.0 & 35.0 & \graynum{21.0}
& 28.0 & 29.0 & 27.0 & \graynum{21.0}
& 67.0 & 55.8 & 27.0 & \graynum{22.5} & \graynum{24.0} & \graynum{22.0}
& \graynum{48.0} & \graynum{42.2} & \graynum{25.0} & \graynum{19.5} & \graynum{25.0} & 25.8 \\

\cellcolor{purple!6}Gemma-3-4B & 18 & \graynum{28.5}
& \best{38.0} & \graynum{19.0} & \graynum{25.0} & \graynum{21.0}
& \graynum{20.0} & \graynum{25.0} & \graynum{24.0} & 26.0
& \graynum{35.0} & \graynum{34.0} & \graynum{24.0} & \graynum{23.2} & \graynum{22.0} & \graynum{21.2}
& 56.0 & \graynum{45.8} & 28.0 & 27.8 & 30.0 & \graynum{24.5} \\
\end{tabular}
\end{adjustbox}

\caption{Results on \textsc{\textbf{SpaceNum}} benchmark.
Accuracy (\%) is reported under two major categories: \textit{Dynamic Transition} and \textit{Static Layout}. 
Each category contains both \textsc{Num2Space} and \textsc{Space2Num}.
Avg. denotes the macro-average. \textbf{Bold} and \underline{underline} denote best and second best, and \graynum{gray values} indicate performances that even below random guess.}
\label{tab:main}
\end{table*}

\section{Experiments}

\paragraph{Experimental Setup.}
We evaluate 18 VLMs from 6 model families on \textsc{SpaceNum}, ranging from 2B to 72B~\citep{Qwen2.5-VL, qwen3technicalreport, wang2025internvl3, lu2025ovis2, cosmos_reason2, gemma3}. 
All models are evaluated with the same prompt format, where they are instructed to directly output the option letter without explanations or intermediate reasoning. 
We run inference in bfloat16 precision with Flash Attention 2 for efficient evaluation, with temperature to 0.7, top-p to 0.9, top-k to 50. All experiments are run on 4 NVIDIA H100 (80GB) GPUs.

\subsection{Overall Results}
\label{sec: overall}

\paragraph{Do VLMs possess spatial numerical understanding?}
As shown in Table~\ref{tab:main}, current VLMs struggle to genuinely understand numerical values in spatial settings. Their performance remains close to random guess (30.0\%), with the best model reaching only 39.8\% on average, and several models even falling below the random baseline. These results suggest that \textbf{current models only capture shallow spatial-number correlations instead of truly grounding numerical values in spatial meaning}.

\paragraph{What patterns emerge across different spatial scenarios?}
Dynamic transitions and static layouts exhibit fundamentally different difficulty structures. In dynamic transitions, performance remains consistently low across all action types, with strong models achieving only around 40.0\%, just 10 points above the random baseline (30.0\%). Models show little preference or specialization across actions, suggesting \textbf{a broad failure to model transition dynamics}. In contrast, static layouts exhibit much clearer structural patterns: models perform relatively well in simpler settings such as 1D layouts and desk-scale scenes, but degrade substantially in higher-dimensional and room-scale settings, often only marginally above the 25.0\% random baseline. This suggests that \textbf{layout reasoning difficulty grows systematically with spatial complexity and scene scale}.

\paragraph{How does spatial numerical mapping differ across scenarios?}
The preferred mapping direction differs substantially across scenarios. In dynamic transitions, models consistently perform better in \textsc{Space2Num} than in \textsc{Num2Space}, suggesting that \textbf{dynamic transitions are more vision-dependent}: models benefit from observing spatial changes directly, but struggle to predict future visual outcomes from numerical actions alone. In contrast, static layouts show the opposite trend, where \textsc{Num2Space} consistently outperforms \textsc{Space2Num}. This suggests that \textbf{static layouts rely more on language-side spatial priors}, where models can project numerical structures into space more easily than recovering structured numerical representations from visual scenes.

\begin{figure}[t]
  \centering
  \begin{subfigure}[t]{0.62\linewidth}
    \centering
    \includegraphics[width=\linewidth]{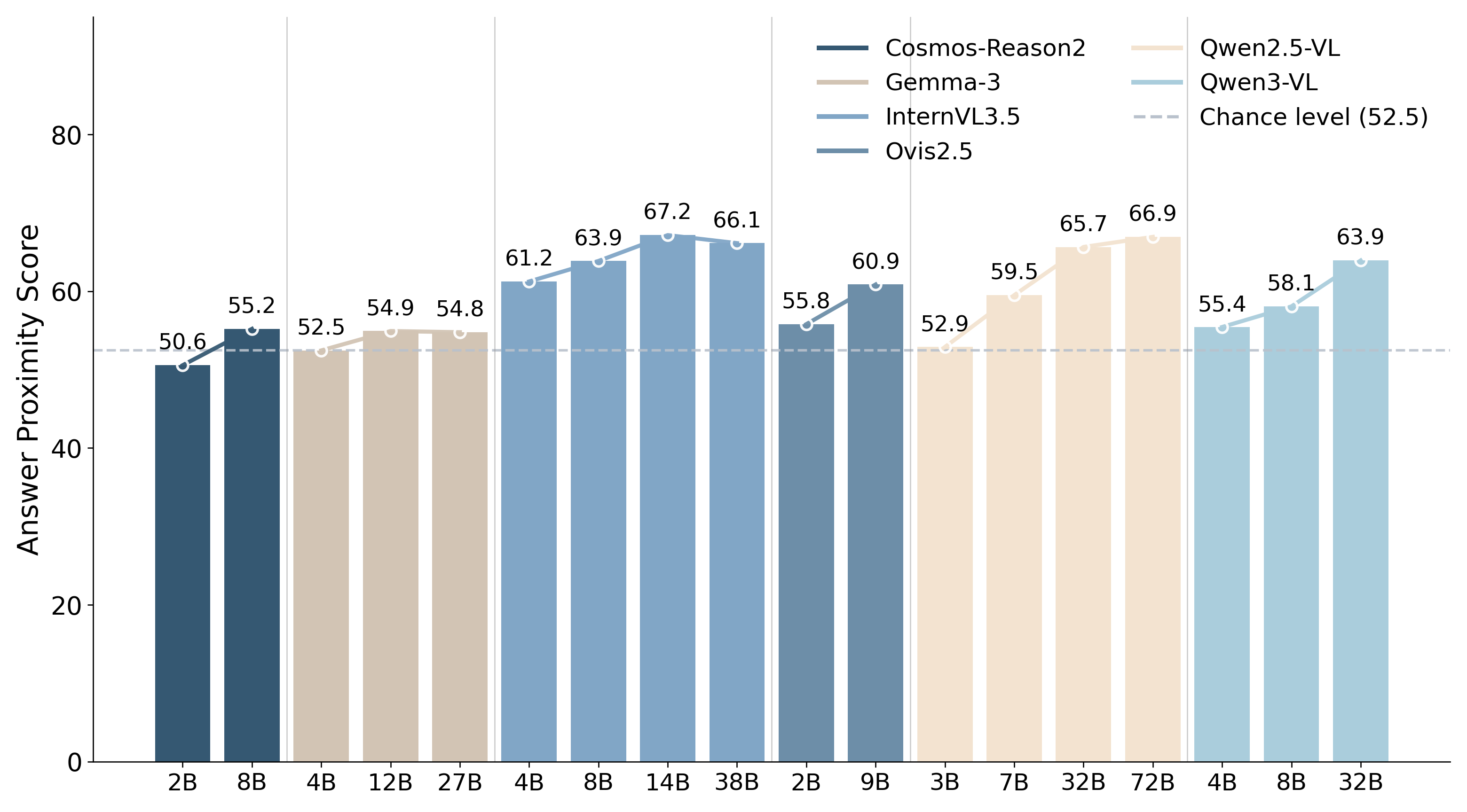}
    \caption{Error proximity in dynamic transitions.}
    \label{fig:err_pattern_dynamic}
  \end{subfigure}
  \hfill
  \begin{subfigure}[t]{0.36\linewidth}
    \centering
    \includegraphics[width=\linewidth]{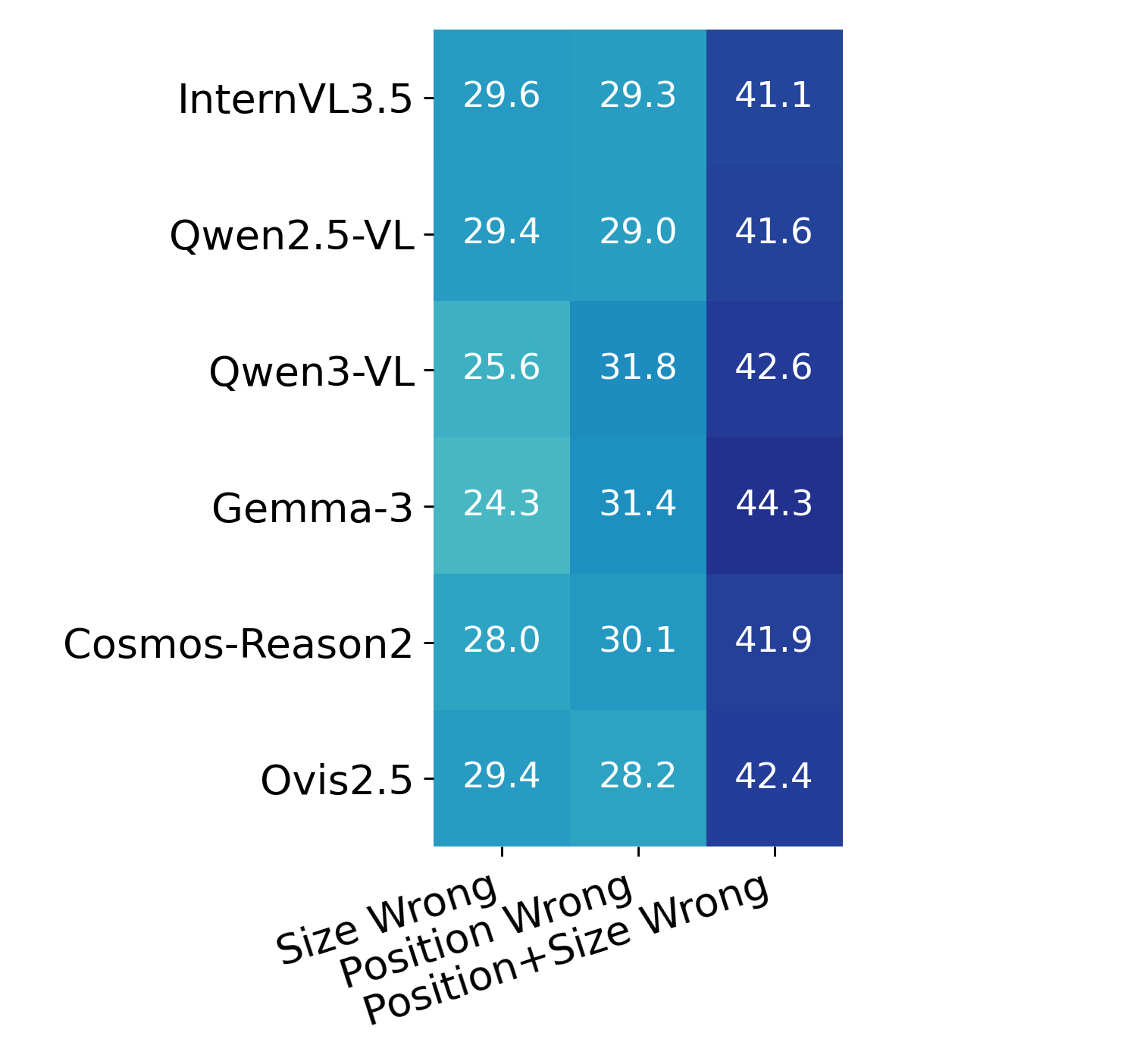}
    \caption{Error decomposition in static layouts.}
    \label{fig:err_pattern_static}
  \end{subfigure}
  \caption{
  Structured analysis of model errors across spatial scenarios.
  Left: larger models tend to make numerically closer mistakes in dynamic transitions.
  Right: static layout failures are dominated by coupled position-and-size errors rather than isolated attribute errors.
  }
  \label{fig:err_pattern_combined}
\end{figure}

\subsection{Structured Analysis of Output Patterns}
\label{sec: structured analysis}

\paragraph{Are larger models making better mistakes?}
Beyond standard multiple-choice accuracy, \textsc{SpaceNum} enables a more structured analysis of model behavior by leveraging the semantic relations among answer choices in different spatial scenarios. For dynamic transitions, we analyze not only exact-match accuracy but also the semantic proximity between the selected answer and the ground truth. Specifically, we assign scores of \{100, 70, 40, 0\} to exact, near, moderate, and far errors according to the numerical distance between the predicted and correct transition magnitudes. Figure~\ref{fig:err_pattern_dynamic} shows a clear trend: as model size increases, predictions become progressively closer to the correct answer even when exact-match accuracy changes only slightly. \textbf{Larger models make less severe transition errors}, suggesting that scaling improves coarse spatial sensitivity even when precise numerical grounding remains difficult.

\paragraph{Do spatial errors decompose across attributes?}
For static layouts, we categorize errors according to whether the predicted layout contains incorrect position, incorrect size, or both. Surprisingly, models consistently favor joint position-and-size errors over single-factor errors across model families, as shown in Figure~\ref{fig:err_pattern_static}. \textbf{Static layout failures are strongly coupled across spatial attributes}: once models fail to establish a coherent layout, errors tend to propagate jointly across position and scale rather than remain isolated. This suggests that current VLMs rely more on coarse holistic matching than disentangled spatial reasoning.

\subsection{Does Reasoning Help Spatial Numerical Understanding?}
\label{sec: reasoning}

To answer this question, we compare reasoning-enabled (\textit{think}) and standard (\textit{non-think}) inference across InternVL3.5-4B/8B/14B and Qwen3-VL-4B/8B/32B. Surprisingly, \textbf{enabling reasoning produces only marginal changes on \textsc{SpaceNum}, with performance differences typically remaining within 1\%.} This suggests that simply generating longer reasoning traces does not substantially improve spatial numerical understanding. We therefore further analyze model traces and identify several recurring failure patterns that explain why reasoning often fails.

\paragraph{Models stop at coarse spatial cues instead of performing fine-grained comparison.}
A common failure is that models identify a plausible spatial cue and terminate reasoning too early. For example, in dynamic transition tasks, a model may observe that ``a new wooden sculpture becomes visible on the left'' and immediately select the corresponding candidate. However, the correct solution requires one more step: comparing how far objects shift across candidates to determine the correct transition magnitude. Similarly, in static layout tasks, models often correctly identify cues such as ``the sofa is left of the tree,'' but fail to compare object size across candidates. In both settings, the model performs coarse cue matching but misses the finer comparison needed to disambiguate similar options.

\paragraph{Models fail to reason counterfactually about motion magnitude.}
Successful \textsc{Space2Num} reasoning often depends on counterfactual magnitude comparison. Correct traces do not only check what changed, but also whether the observed change is large enough to support a candidate magnitude. For example, when estimating a small rotation, correct models explicitly reason that ``most objects remain aligned across the two views,'' and therefore ``a 70$^\circ$ rotation would produce much larger layout changes.'' In contrast, incorrect traces often map any noticeable visual change directly to a large number, e.g., ``the perspective changes noticeably, suggesting a large right rotation.'' These traces focus only on changed evidence while ignoring stable evidence.

\paragraph{Models reason in image space instead of the defined coordinate system.}
Another recurring failure is that models rely on generic image-space priors rather than constructing the coordinate system defined by the anchor objects. For instance, some traces directly map ``left in the image'' to a smaller $x$ value, reasoning that ``the piano is positioned on the left side of the image, so it should have a smaller x-coordinate.'' However, the correct solution requires first establishing the coordinate frame using the provided anchors and then reasoning relative to that frame. Similarly, models may correctly describe an object as ``behind'' another object but still assign the wrong depth direction because they fail to align the scene with the task-defined coordinate system.

\begin{figure}[t]
  \centering

  \begin{subfigure}[t]{0.50\linewidth}
    \centering
    \includegraphics[width=\linewidth]{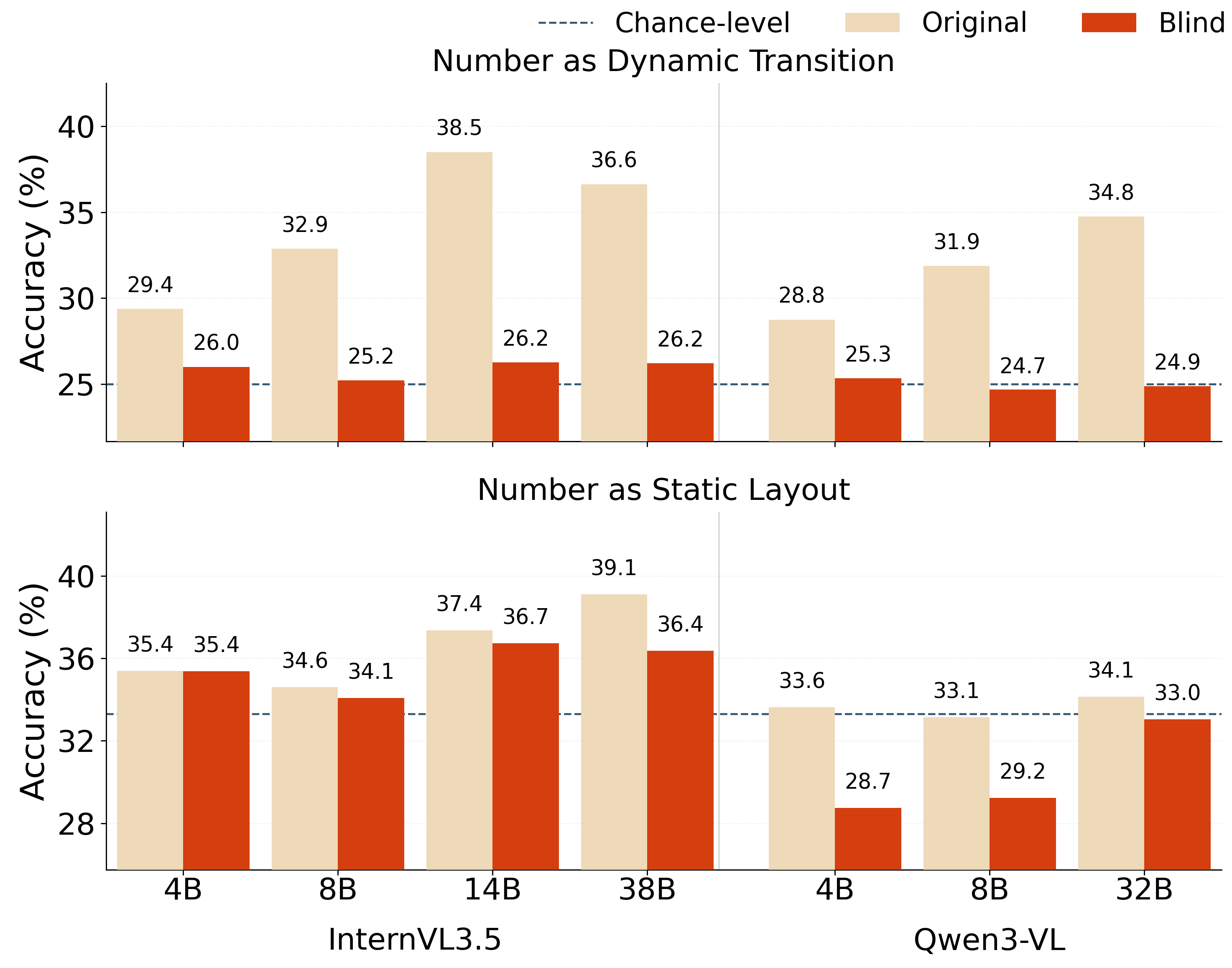}
    \caption{Blind testing.}
    \label{fig:blind}
  \end{subfigure}
  \hfill
  \begin{subfigure}[t]{0.48\linewidth}
    \centering
    \includegraphics[width=\linewidth]{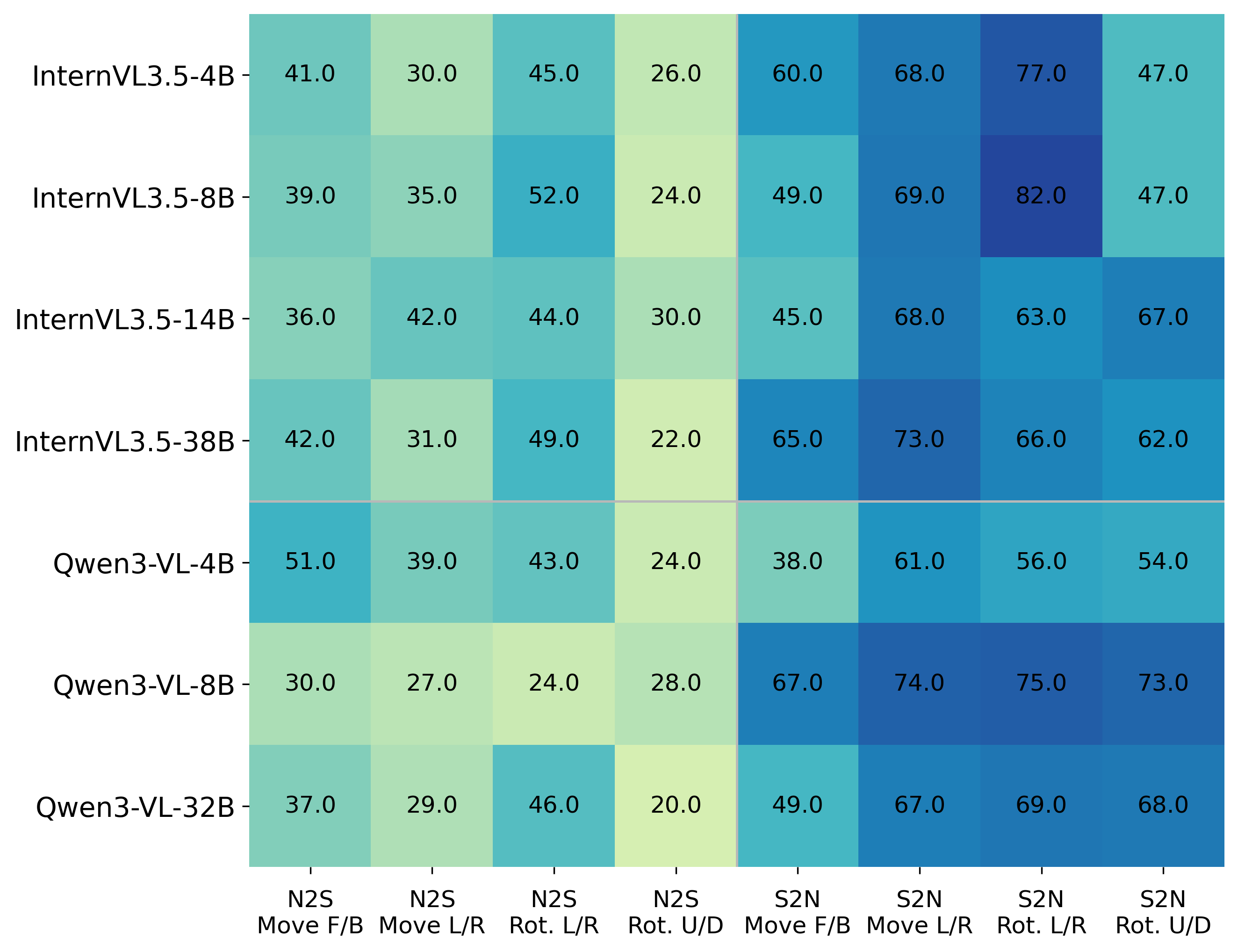}
    \caption{Per-action mapping asymmetry.}
    \label{fig:action_symm}
  \end{subfigure}

  \begin{subfigure}[t]{0.99\linewidth}
    \centering
    \includegraphics[width=\linewidth]{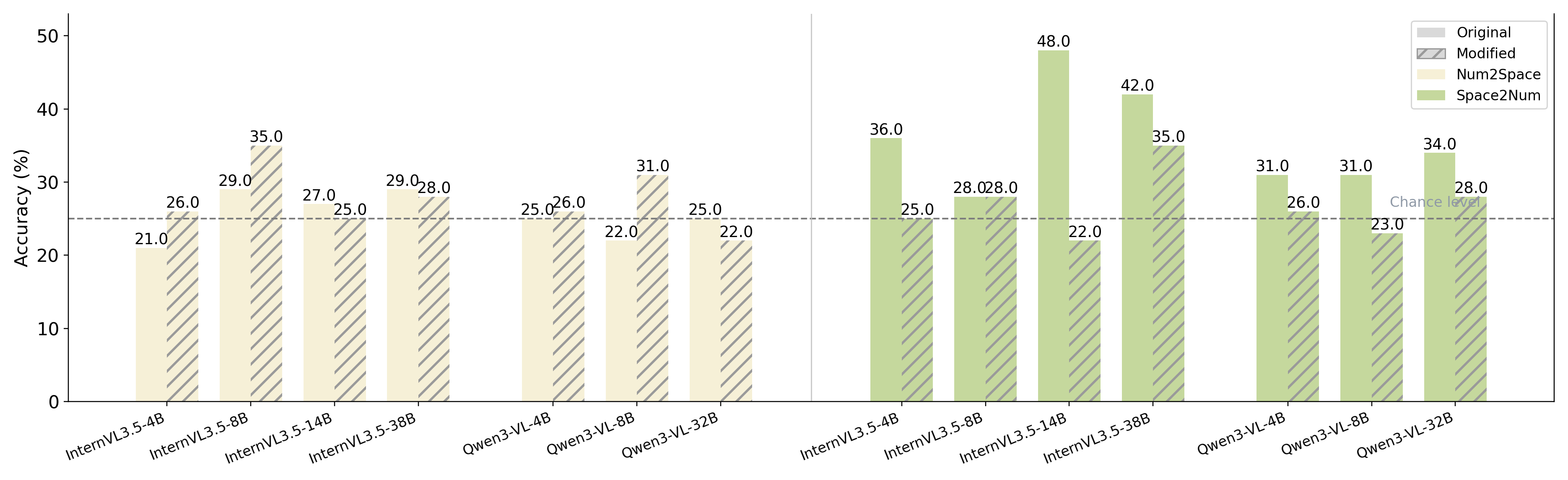}
    \caption{Rotational symmetry analysis.}
    \label{fig:rotation_symm}
  \end{subfigure}

  \caption{
  Additional analyses under dynamic transitions.
  Top left: blind testing by masking visual inputs.
  Top right: per-action comparison between \textsc{Num2Space} and \textsc{Space2Num}.
  Bottom: rotational symmetry analysis under equivalent transformations.
  }
  \label{fig:dynamic_transition_analysis}
\end{figure}

\subsection{Modality Asymmetry in Spatial Numerical Understanding}
\label{sec: modality}

\paragraph{How much do models rely on visual information?}
To examine whether models truly depend on visual grounding, we conduct a blind testing study by replacing images with fully black inputs while keeping the task format unchanged. As shown in Figure~\ref{fig:blind}, masking visual inputs causes a substantial performance drop for \textit{number as dynamic transition}, while the effect is much smaller for \textit{number as static layout}. \textbf{Dynamic transitions are significantly more vision-dependent}, whereas static layouts can often be partially solved through language-side priors or shortcut patterns without fully grounding the visual scene.

\paragraph{Is spatial numerical mapping balanced across actions?}
We further compare \textsc{Num2Space} and \textsc{Space2Num} at the level of individual actions. Figure~\ref{fig:action_symm} shows that, for almost every action type, \textit{Space2Num} consistently outperforms \textit{Num2Space}. \textbf{The asymmetry between the two mapping directions persists even under the same underlying action dynamics}, suggesting that models are systematically better at grounding numbers from observed visual changes than predicting future visual outcomes from numerical actions.

\paragraph{Do models learn geometrically consistent spatial mappings?}
Finally, we probe \textit{Space2Num} under rotational symmetry transformations. Ideally, equivalent actions such as rotating left by $20^\circ$ and rotating right by $340^\circ$ should lead to consistent numerical predictions. However, Figure~\ref{fig:rotation_symm} shows substantial performance drops under these symmetric transformations. \textbf{The mapping from vision to numbers lacks geometric consistency and invariance}, suggesting that models fail to build stable numerical representations from visual spatial changes.

\begin{figure*}[t]
\centering

\begin{minipage}[t]{0.32\textwidth}
\vspace{0pt}
\centering
\includegraphics[width=\linewidth]{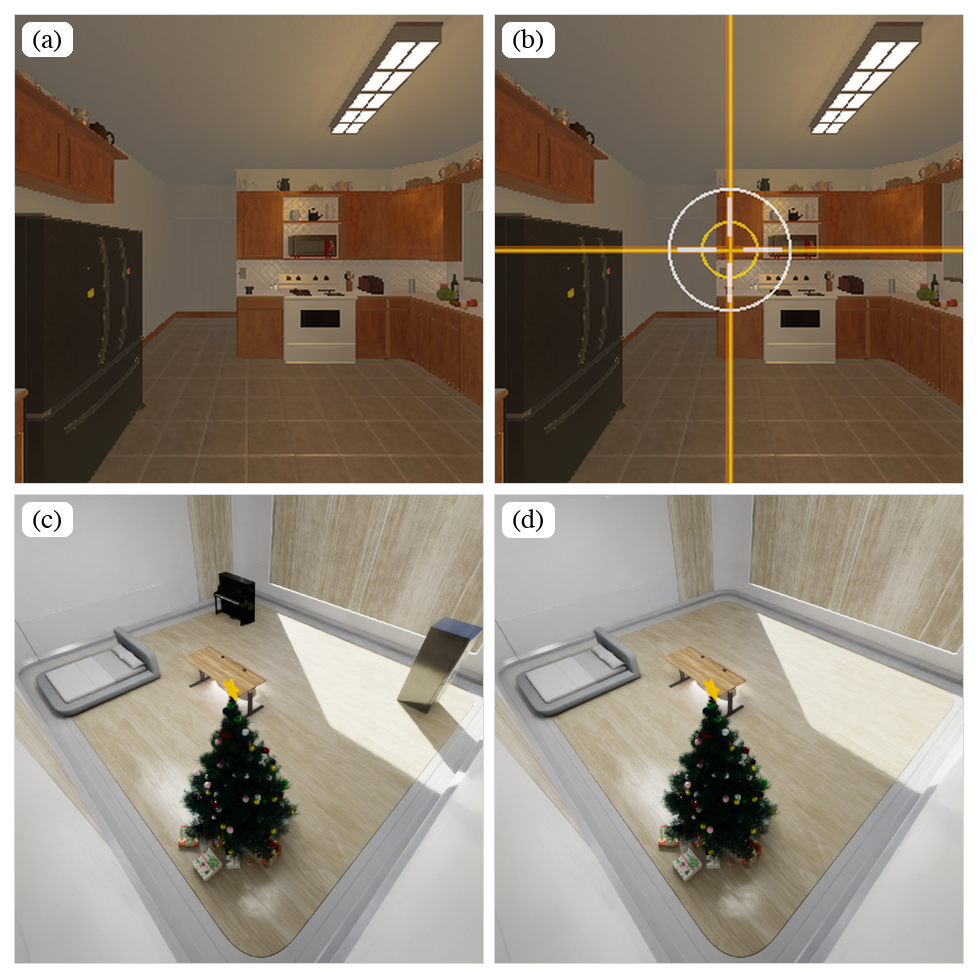}
\end{minipage}
\hfill
\begin{minipage}[t]{0.60\textwidth}
\vspace{6pt}
\centering
\small
\setlength{\tabcolsep}{5pt}
\renewcommand{\arraystretch}{1.1}
\begin{tabular}{lcc}
\toprule
\textbf{Model} 
& \shortstack{\textbf{Add Anchor}\\\textbf{(Transition)}} 
& \shortstack{\textbf{Reduce Objects}\\\textbf{(Layout)}} \\
\midrule
InternVL3.5-4B  & -0.3\% & -1.6\% \\
InternVL3.5-8B  & -1.3\% & -0.0\% \\
InternVL3.5-14B & -2.3\% & -0.6\% \\
InternVL3.5-38B & +0.9\% & +0.1\% \\
\midrule
Qwen3-VL-4B     & +0.5\% & -1.1\% \\
Qwen3-VL-8B     & -2.5\% & -0.1\% \\
Qwen3-VL-32B    & -1.0\% & -0.3\% \\
\bottomrule
\end{tabular}
\end{minipage}

\caption{
Visual-side interventions.
Left: adding anchors for dynamic transitions and reducing objects for static layouts.
Right: both interventions lead to only minor and inconsistent performance changes.
}
\label{fig:focus_factor}
\end{figure*}

\begin{figure*}[t]
  \centering

  \begin{minipage}[t]{0.49\textwidth}
    \vspace{14pt}
    \centering
    \small
    \resizebox{\linewidth}{!}{
    \begin{tabular}{lccc}
    \toprule
    \textbf{Model} 
    & \shortstack{\textbf{NL $\Delta$}\\\textbf{(Trans.)}} 
    & \shortstack{\textbf{Int. $\Delta$}\\\textbf{(Trans.)}} 
    & \shortstack{\textbf{Int. $\Delta$}\\\textbf{(Layout)}} \\
    \midrule
    InternVL3.5-4B  & -0.3\% & +0.7\% & -0.9\% \\
    InternVL3.5-8B  & -1.0\% & +0.8\% & +1.6\% \\
    InternVL3.5-14B & -1.3\% & -0.6\% & -1.3\% \\
    InternVL3.5-38B & -1.8\% & +4.5\% & -1.8\% \\
    \midrule
    Qwen3-VL-4B     & -1.2\% & -1.7\% & +0.2\% \\
    Qwen3-VL-8B     & +2.0\% & -2.0\% & +0.5\% \\
    Qwen3-VL-32B    & -1.2\% & +3.3\% & +1.3\% \\
    \bottomrule
    \end{tabular}
    }
    \caption*{(a) Numerical representation changes.}
  \end{minipage}
  \hfill
  \begin{minipage}[t]{0.49\textwidth}
    \vspace{0pt}
    \centering
    \includegraphics[width=\linewidth]{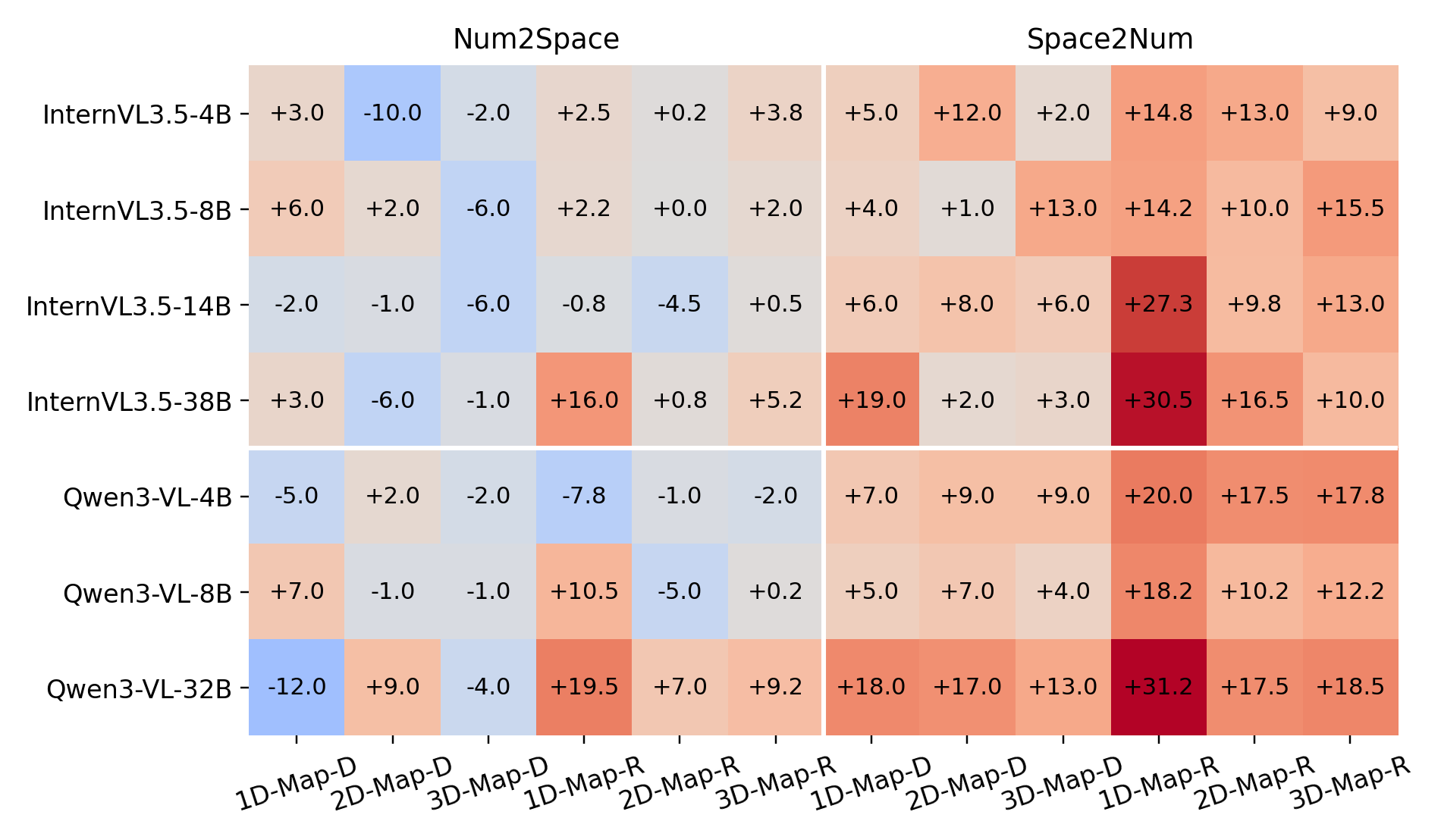}
    \caption*{(b) Visual abstraction for layouts.}
  \end{minipage}

  \caption{
  Representation-side interventions.
  Left: changing numerical representations in dynamic transitions and layouts.
  Right: simplifying layouts into structured visual abstractions.
  }
  \label{fig:factor_analysis}
\end{figure*}

\subsection{Disentangling Factors in Spatial Numerical Understanding}
\label{sec: factors}

\paragraph{Can simple visual interventions improve spatial grounding?}
We first modify visual inputs in both scenarios. For dynamic transitions, we add explicit visual anchors to help models measure spatial changes. For static layouts, we reduce irrelevant objects to simplify visual grounding. However, Figure~\ref{fig:focus_factor} shows that both interventions lead to only minor and inconsistent improvements. \textbf{The core limitation is not caused by missing visual references or cluttered scenes.}

\paragraph{Does the numerical representation itself matter?}
We then vary how numerical values are expressed. Converting numbers into natural language yields negligible gains, while integer-scaled representations (e.g., meters to centimeters) provide only limited improvements for larger models in transition tasks. As shown in Figure~\ref{fig:factor_analysis}(a), performance in layout reasoning remains largely unchanged. \textbf{The bottleneck does not primarily lie in the surface form of numerical representations.}

\paragraph{Do models struggle to abstract spatial structure from images?}
Since neither visual simplification nor numerical reformulation resolves the issue, we further investigate whether models fail to extract structured spatial representations from raw images. We therefore replace layout images with progressively more structured abstractions, including points, 2D boxes, and 3D boxes. Figure~\ref{fig:factor_analysis}(b) shows that this substantially improves \textsc{Space2Num}, while providing less effects for \textsc{Num2Space}. \textbf{The main bottleneck lies in vision-to-structure abstraction}: current VLMs struggle to transform raw visual observations into structured spatial representations suitable for numerical reasoning.

\subsection{Tuning Spatial Numerical Understanding}
\label{sec: tuning}

\begin{figure*}[t]
\centering

\begin{minipage}[t]{0.48\textwidth}
\vspace{3pt}
\centering
\includegraphics[width=\linewidth]{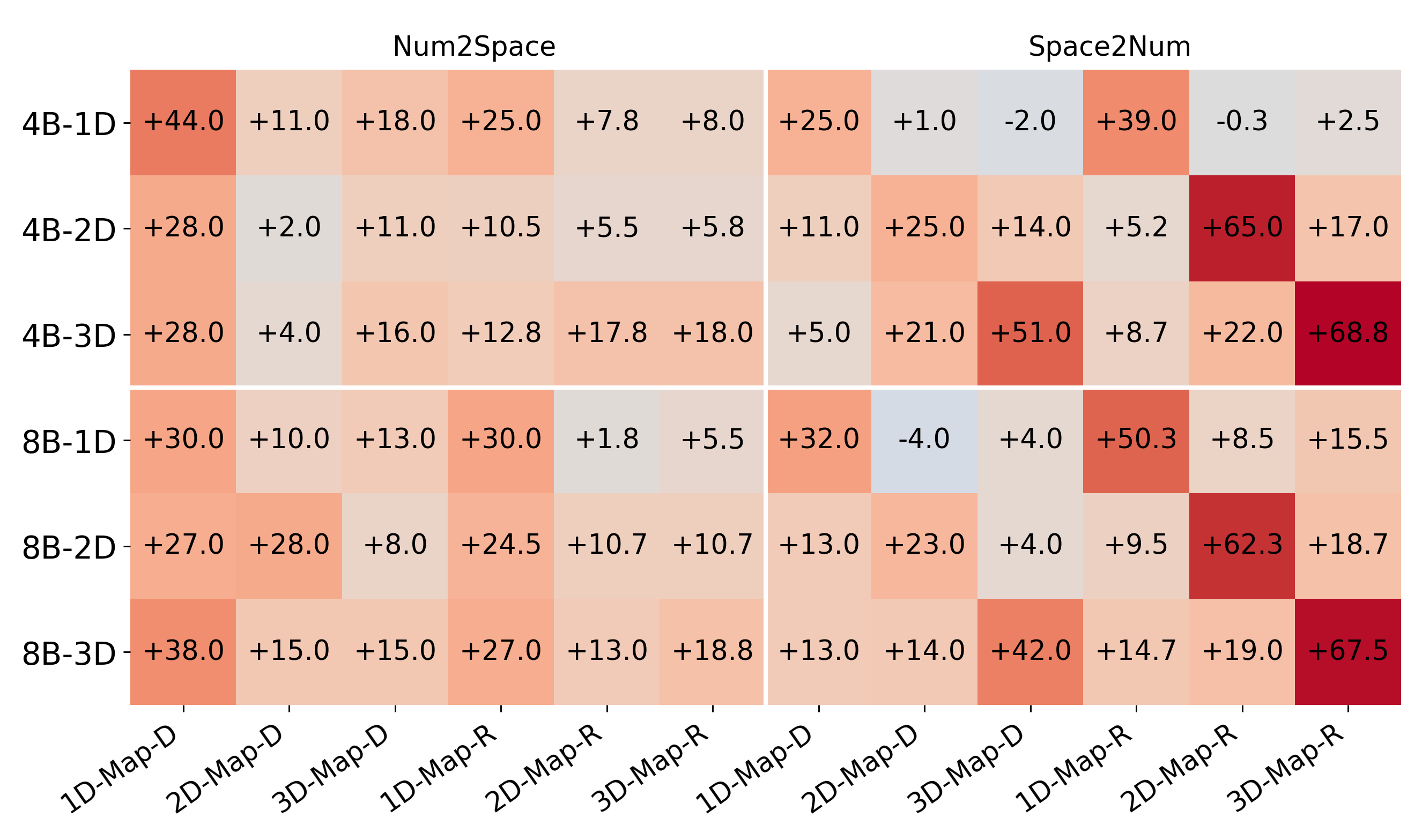}
\caption*{(a) Cross-dimension tuning transfer.}
\end{minipage}
\hfill
\begin{minipage}[t]{0.48\textwidth}
\vspace{0pt}
\centering
\includegraphics[width=\linewidth]{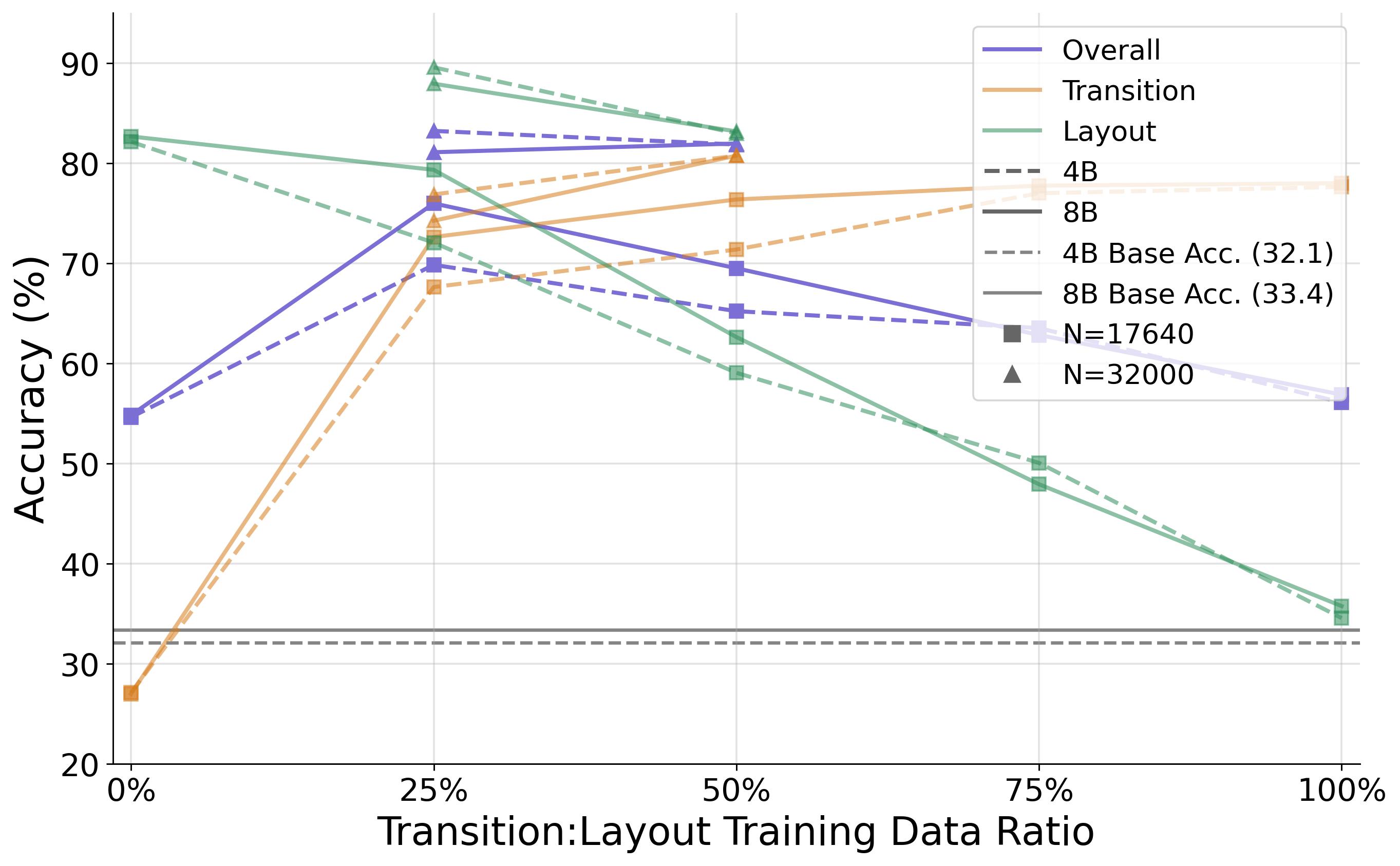}
\caption*{(b) Training data mixture and scaling.}
\end{minipage}

\caption{
Tuning analysis for spatial numerical understanding.
Left: transfer patterns across different spatial dimensions.
Right: effects of data mixture ratios and training scale.
}
\label{fig:tuning_analysis}
\end{figure*}


\paragraph{Can spatial reasoning transfer across dimensions?}
We fine-tune Qwen3-VL-4B and Qwen3-VL-8B with LoRA using a learning rate of \(1\times10^{-4}\), cosine decay with a 0.1 warmup ratio, bfloat16 precision, a maximum sequence length of 2048, LoRA rank 8 and alpha 16, and an effective batch size of 128 for 3 epochs. Figure~\ref{fig:tuning_analysis}(a) shows a clear diagonal pattern: tuning on a particular dimension yields the largest improvement on the same dimension, suggesting that different dimensions encode distinct spatial structures. At the same time, tuning on 1D data also improves performance on 2D and 3D settings, especially for larger models and more clearly in \textsc{Num2Space}. \textbf{Lower-dimensional spatial reasoning can partially transfer to higher-dimensional settings}, although the transfer remains limited.

\paragraph{What data recipe leads to the best spatial reasoning ability?}
We next vary the ratio between transition and layout data. As shown in Figure~\ref{fig:tuning_analysis}(b), the best overall performance consistently emerges when transition data accounts for roughly 25\% and layout data accounts for roughly 75\%. Increasing the total amount of training data further improves performance under the same ratio. \textbf{Both data composition and training scale substantially affect spatial numerical understanding}, with layout-heavy mixtures producing the strongest overall capability.

\begin{wraptable}{r}{0.43\linewidth}
\vspace{-0.4cm}
\centering
\small
\setlength{\tabcolsep}{4pt}
\renewcommand{\arraystretch}{1.05}
\begin{tabular}{lcc}
\toprule
\textbf{Metric} & \textbf{Strict} & \textbf{Graded} \\
\midrule
Transition    & +6.38 & \textbf{+6.88} \\
Layout        & +6.64 & \textbf{+7.60} \\
\midrule
Num2Space     & +8.10 & \textbf{+9.37} \\
Space2Num     & +5.05 & \textbf{+5.52} \\
\bottomrule
\end{tabular}
\vspace{-0.2cm}
\caption{
Performance improvement under two different reward designs.
}
\label{tab:reward_design}
\vspace{-0.3cm}
\end{wraptable}

\paragraph{Does RL help, and does reward design matter?}
We further study RL tuning on the 4B model using GRPO with LoRA rank 64 and alpha 64, a learning rate of \(1\times10^{-5}\), rollout batch size 128, actor batch size 64, and 5 rollouts per prompt. We compare a strict exact-match reward and a graded reward based on error magnitude. As shown in Table~\ref{tab:reward_design}, RL brings only limited gains overall, while graded rewards perform slightly better than strict rewards.

\begin{wraptable}[6]{r}{0.38\linewidth}
\centering
\small
\setlength{\tabcolsep}{5pt}
\begin{tabular}{lcc}
\toprule
\textbf{Metric} & \textbf{4B $\Delta$} & \textbf{8B $\Delta$} \\
\midrule
OS-Motion & +5.5 & +4.5 \\
SAT-AC    & +8.1 & +18.9 \\
SAT-OM    & +34.8 & +43.5 \\
\bottomrule
\end{tabular}
\vspace{-0.2cm}
\caption{
Transfered performance.
}
\label{tab:ext_transfer}
\vspace{-0.3cm}
\end{wraptable}
\paragraph{Does the learned ability generalize beyond \textsc{SpaceNum}?}
Finally, we evaluate tuned models on external spatial reasoning benchmarks. Table~\ref{tab:ext_transfer} shows consistent improvements across all tasks. 
Gains on OmniSpatial Motion~\citep{jia2025omnispatial} indicate better understanding of camera movement, while improvements on SAT Action Consequence and Object Movement~\citep{ray2024sat} demonstrate stronger reasoning about action outcomes and object dynamics. The improvements are particularly large for the 8B model. \textbf{The learned capability transfers beyond our benchmark}, suggesting that tuning improves general spatial reasoning ability rather than merely overfitting in our settings.

\section{Related Works}

\paragraph{Spatial reasoning in dynamic and embodied environments.}
Recent works study whether VLMs can reason about spatial changes caused by actions, motion, and embodied interactions. SAT evaluates dynamic spatial aptitude through action consequence prediction, object movement, perspective taking, and spatial aiming tasks~\citep{ray2024sat}. OmniSpatial provides a comprehensive benchmark for spatial reasoning over camera motion, object motion, perspective transformation, and interaction-centered scenarios~\citep{jia2025omnispatial}. VSI-Bench evaluates whether MLLMs can see, remember, and recall spatial environments from sequential visual observations~\citep{yang2025thinking}. MVoT improves spatial reasoning by encouraging models to imagine intermediate visual states during reasoning~\citep{li2025imagine}. SpaceTools studies tool-augmented spatial reasoning through interactive reinforcement learning with external spatial tools~\citep{chen2025spacetools}. These works show that current VLMs struggle with dynamic spatial reasoning and spatial transformations. However, they mainly focus on whether models understand spatial changes themselves, rather than whether the numerical values parameterizing these transitions are truly grounded in spatial meaning.

\paragraph{Spatial understanding and structured spatial reasoning.}
Another line of work studies whether VLMs can infer spatial relations, metric structure, and 3D layouts from visual observations. Early benchmarks evaluate relations such as left/right, above/below, and object-centric configurations, showing that VLMs often struggle with spatial prepositions despite strong object recognition ability~\citep{liu2023visual,kamath2023s,rajabi2024gsr,du2024embspatial}. More recent works extend this evaluation to metric reasoning, geometric reasoning, open-space understanding, and domain-specific 3D reasoning~\citep{daxberger2025mm,zhang2025open3dvqa,wu2025spatialscore,xu2026earthspatialbench}. Beyond evaluation, several works inject explicit spatial structures into VLMs through spatial annotations, region-level grounding, coordinates, distances, layouts, and 3D priors~\citep{chen2024spatialvlm,cheng2024spatialrgpt,ma2024spatialpin,liao2024qspatial,daxberger2025mm,huang2025video2layout}. More recently, SpatialReasoner studies explicit and generalizable 3D spatial reasoning through structured spatial representations~\citep{ma2025spatialreasoner}. Together, these works improve structured spatial understanding and reasoning ability in VLMs, but they mainly treat numbers as auxiliary labels or outputs, rather than directly studying whether numerical values themselves are grounded as meaningful spatial quantities.

In contrast to prior work, \textsc{SpaceNum} directly studies spatial numerical understanding: whether VLMs can ground numerical values as meaningful spatial quantities across both dynamic transitions and static layouts. Beyond benchmark evaluation, we further analyze the asymmetry, failure patterns, reasoning behaviors, and tuning characteristics of spatial numerical grounding in current VLMs.

\section{Conclusion}

In this work, we study whether current Vision Language Models (VLMs) truly understand numerical values in spatial settings through \textsc{SpaceNum}, a unified benchmark covering both dynamic transitions and static layouts. Our experiments show that current VLMs largely fail to ground numbers in spatial meaning, often relying on shallow spatial cues instead of stable spatial reasoning. Through systematic analyses, we further show that these failures arise from weak spatial abstraction, asymmetric vision-number mappings, and the inability to build structured coordinate-aware representations. Although tuning partially improves performance and transfers to related benchmarks, substantial gaps still remain. We hope \textsc{SpaceNum} can serve as a useful benchmark and diagnostic framework for future research on spatial numerical understanding in VLMs.

\paragraph{Limitations and future work.}
Our study mainly focuses on controlled spatial settings with discrete candidate-based evaluation and simulated environments. Extending spatial numerical understanding to more open-ended real-world scenes, embodied interactions, and continuous spatial prediction settings remains an important direction for future work. We also mainly analyze failures from the vision and language sides, while how VLMs internally perform spatial reasoning remains largely unexplored. Although we conduct preliminary attention-based analyses, severe attention collapse in current VLMs makes it difficult to obtain clear conclusions. Understanding the internal mechanisms behind spatial numerical reasoning therefore remains an important future direction.

\bibliographystyle{plainnat}
\bibliography{references}

@article{kolve2017ai2,
  title={Ai2-thor: An interactive 3d environment for visual ai},
  author={Kolve, Eric and Mottaghi, Roozbeh and Han, Winson and VanderBilt, Eli and Weihs, Luca and Herrasti, Alvaro and Deitke, Matt and Ehsani, Kiana and Gordon, Daniel and Zhu, Yuke and others},
  journal={arXiv preprint arXiv:1712.05474},
  year={2017}
}

@misc{nvidia_isaac_sim,
  title = {NVIDIA Isaac Sim},
  author = {{NVIDIA Corporation}},
  year = {2023},
  howpublished = {\url{https://developer.nvidia.com/isaac-sim}}
}

@misc{blenderkit,
  title = {BlenderKit: Online Asset Library for Blender},
  author = {{BlenderKit}},
  year = {2023},
  howpublished = {\url{https://www.blenderkit.com/}}
}

@article{Qwen2.5-VL,
  title={Qwen2.5-VL Technical Report},
  author={Bai, Shuai and Chen, Keqin and Liu, Xuejing and Wang, Jialin and Ge, Wenbin and Song, Sibo and Dang, Kai and Wang, Peng and Wang, Shijie and Tang, Jun and Zhong, Humen and Zhu, Yuanzhi and Yang, Mingkun and Li, Zhaohai and Wan, Jianqiang and Wang, Pengfei and Ding, Wei and Fu, Zheren and Xu, Yiheng and Ye, Jiabo and Zhang, Xi and Xie, Tianbao and Cheng, Zesen and Zhang, Hang and Yang, Zhibo and Xu, Haiyang and Lin, Junyang},
  journal={arXiv preprint arXiv:2502.13923},
  year={2025}
}

@misc{qwen3technicalreport,
      title={Qwen3 Technical Report}, 
      author={Qwen Team},
      year={2025},
      eprint={2505.09388},
      archivePrefix={arXiv},
      primaryClass={cs.CL},
      url={https://arxiv.org/abs/2505.09388}, 
}

@article{wang2025internvl3,
  title={Internvl3. 5: Advancing open-source multimodal models in versatility, reasoning, and efficiency},
  author={Wang, Weiyun and Gao, Zhangwei and Gu, Lixin and Pu, Hengjun and Cui, Long and Wei, Xingguang and Liu, Zhaoyang and Jing, Linglin and Ye, Shenglong and Shao, Jie and others},
  journal={arXiv preprint arXiv:2508.18265},
  year={2025}
}

@article{lu2025ovis2,
  title={Ovis2. 5 technical report},
  author={Lu, Shiyin and Li, Yang and Xia, Yu and Hu, Yuwei and Zhao, Shanshan and Ma, Yanqing and Wei, Zhichao and Li, Yinglun and Duan, Lunhao and Zhao, Jianshan and others},
  journal={arXiv preprint arXiv:2508.11737},
  year={2025}
}

@misc{cosmos_reason2,
  title = {Cosmos-Reason2: Open Reasoning Vision-Language Models for Physical AI},
  author = {NVIDIA},
  year = {2026},
  howpublished = {\url{https://huggingface.co/collections/nvidia/cosmos-reason2}},
  note = {Accessed: 2026-05-01}
}

@misc{gemma3,
  title = {Gemma 3},
  author = {Google DeepMind},
  year = {2025},
  howpublished = {\url{https://deepmind.google/models/gemma/gemma-3/}},
  note = {Accessed: 2026-05-01}
}

@article{ray2024sat,
  title={Sat: Dynamic spatial aptitude training for multimodal language models},
  author={Ray, Arijit and Duan, Jiafei and Brown, Ellis and Tan, Reuben and Bashkirova, Dina and Hendrix, Rose and Ehsani, Kiana and Kembhavi, Aniruddha and Plummer, Bryan A and Krishna, Ranjay and others},
  journal={arXiv preprint arXiv:2412.07755},
  year={2024}
}

@article{jia2025omnispatial,
  title={Omnispatial: Towards comprehensive spatial reasoning benchmark for vision language models},
  author={Jia, Mengdi and Qi, Zekun and Zhang, Shaochen and Zhang, Wenyao and Yu, Xinqiang and He, Jiawei and Wang, He and Yi, Li},
  journal={arXiv preprint arXiv:2506.03135},
  year={2025}
}

@inproceedings{yang2025thinking,
  title={Thinking in space: How multimodal large language models see, remember, and recall spaces},
  author={Yang, Jihan and Yang, Shusheng and Gupta, Anjali W and Han, Rilyn and Fei-Fei, Li and Xie, Saining},
  booktitle={Proceedings of the Computer Vision and Pattern Recognition Conference},
  pages={10632--10643},
  year={2025}
}

@article{li2025imagine,
  title={Imagine while reasoning in space: Multimodal visualization-of-thought},
  author={Li, Chengzu and Wu, Wenshan and Zhang, Huanyu and Xia, Yan and Mao, Shaoguang and Dong, Li and Vuli{\'c}, Ivan and Wei, Furu},
  journal={arXiv preprint arXiv:2501.07542},
  year={2025}
}

@article{chen2025spacetools,
  title={SpaceTools: Tool-Augmented Spatial Reasoning via Double Interactive RL},
  author={Chen, Siyi and Uy, Mikaela Angelina and Song, Chan Hee and Ladhak, Faisal and Murali, Adithyavairavan and Qu, Qing and Birchfield, Stan and Blukis, Valts and Tremblay, Jonathan},
  journal={arXiv preprint arXiv:2512.04069},
  year={2025}
}

@article{liu2023visual,
  title={Visual spatial reasoning},
  author={Liu, Fangyu and Emerson, Guy and Collier, Nigel},
  journal={Transactions of the Association for Computational Linguistics},
  volume={11},
  pages={635--651},
  year={2023},
  publisher={MIT Press One Broadway, 12th Floor, Cambridge, Massachusetts 02142, USA~…}
}

@inproceedings{kamath2023s,
  title={What’s “up” with vision-language models? investigating their struggle with spatial reasoning},
  author={Kamath, Amita and Hessel, Jack and Chang, Kai-Wei},
  booktitle={Proceedings of the 2023 Conference on Empirical Methods in Natural Language Processing},
  pages={9161--9175},
  year={2023}
}

@article{rajabi2024gsr,
  title={Gsr-bench: A benchmark for grounded spatial reasoning evaluation via multimodal llms},
  author={Rajabi, Navid and Kosecka, Jana},
  journal={arXiv preprint arXiv:2406.13246},
  year={2024}
}

@inproceedings{du2024embspatial,
  title={Embspatial-bench: Benchmarking spatial understanding for embodied tasks with large vision-language models},
  author={Du, Mengfei and Wu, Binhao and Li, Zejun and Huang, Xuan-Jing and Wei, Zhongyu},
  booktitle={Proceedings of the 62nd Annual Meeting of the Association for Computational Linguistics (Volume 2: Short Papers)},
  pages={346--355},
  year={2024}
}

@inproceedings{daxberger2025mm,
  title={Mm-spatial: Exploring 3d spatial understanding in multimodal llms},
  author={Daxberger, Erik and Wenzel, Nina and Griffiths, David and Gang, Haiming and Lazarow, Justin and Kohavi, Gefen and Kang, Kai and Eichner, Marcin and Yang, Yinfei and Dehghan, Afshin and others},
  booktitle={Proceedings of the IEEE/CVF International Conference on Computer Vision},
  pages={7395--7408},
  year={2025}
}

@article{zhang2025open3dvqa,
  title={Open3D-VQA: A Benchmark for Comprehensive Spatial Reasoning with Multimodal Large Language Model in Open Space},
  author={Zhang, Weichen and Zhou, Zile and Zeng, Xin and Liu, Xuchen and Fang, Jianjie and Gao, Chen and Li, Yong and Cui, Jinqiang and Chen, Xinlei and Zhang, Xiao-Ping},
  journal={arXiv preprint arXiv:2503.11094},
  year={2025}
}

@article{wu2025spatialscore,
  title={Spatialscore: Towards unified evaluation for multimodal spatial understanding},
  author={Wu, Haoning and Huang, Xiao and Chen, Yaohui and Zhang, Ya and Wang, Yanfeng and Xie, Weidi},
  journal={arXiv e-prints},
  pages={arXiv--2505},
  year={2025}
}

@article{xu2026earthspatialbench,
  title={EarthSpatialBench: Benchmarking Spatial Reasoning Capabilities of Multimodal LLMs on Earth Imagery},
  author={Xu, Zelin and Zhang, Yupu and Adhikari, Saugat and Islam, Saiful and Xiao, Tingsong and Liu, Zibo and Chen, Shigang and Yan, Da and Jiang, Zhe},
  journal={arXiv preprint arXiv:2602.15918},
  year={2026}
}

@inproceedings{chen2024spatialvlm,
  title={Spatialvlm: Endowing vision-language models with spatial reasoning capabilities},
  author={Chen, Boyuan and Xu, Zhuo and Kirmani, Sean and Ichter, Brain and Sadigh, Dorsa and Guibas, Leonidas and Xia, Fei},
  booktitle={Proceedings of the IEEE/CVF Conference on Computer Vision and Pattern Recognition},
  pages={14455--14465},
  year={2024}
}

@article{cheng2024spatialrgpt,
  title={Spatialrgpt: Grounded spatial reasoning in vision-language models},
  author={Cheng, An-Chieh and Yin, Hongxu and Fu, Yang and Guo, Qiushan and Yang, Ruihan and Kautz, Jan and Wang, Xiaolong and Liu, Sifei},
  journal={Advances in Neural Information Processing Systems},
  volume={37},
  pages={135062--135093},
  year={2024}
}

@article{ma2024spatialpin,
  title={Spatialpin: Enhancing spatial reasoning capabilities of vision-language models through prompting and interacting 3d priors},
  author={Ma, Chenyang and Lu, Kai and Cheng, Ta-Ying and Trigoni, Niki and Markham, Andrew},
  journal={Advances in neural information processing systems},
  volume={37},
  pages={68803--68832},
  year={2024}
}

@inproceedings{liao2024qspatial,
  title={Reasoning paths with reference objects elicit quantitative spatial reasoning in large vision-language models},
  author={Liao, Yuan-Hong and Mahmood, Rafid and Fidler, Sanja and Acuna, David},
  booktitle={Proceedings of the 2024 Conference on Empirical Methods in Natural Language Processing},
  pages={17028--17047},
  year={2024}
}

@article{ma2025spatialreasoner,
  title={Spatialreasoner: Towards explicit and generalizable 3d spatial reasoning},
  author={Ma, Wufei and Chou, Yu-Cheng and Liu, Qihao and Wang, Xingrui and de Melo, Celso and Xie, Jianwen and Yuille, Alan},
  journal={arXiv preprint arXiv:2504.20024},
  year={2025}
}

@article{pi2024image,
  title={Image textualization: An automatic framework for creating accurate and detailed image descriptions},
  author={Pi, Renjie and Zhang, Jianshu and Zhang, Jipeng and Pan, Rui and Chen, Zhekai and Zhang, Tong},
  journal={arXiv preprint arXiv:2406.07502},
  year={2024}
}

@article{dai2023instructblip,
  title={Instructblip: Towards general-purpose vision-language models with instruction tuning},
  author={Dai, Wenliang and Li, Junnan and Li, Dongxu and Tiong, Anthony and Zhao, Junqi and Wang, Weisheng and Li, Boyang and Fung, Pascale N and Hoi, Steven},
  journal={Advances in neural information processing systems},
  volume={36},
  pages={49250--49267},
  year={2023}
}

@article{huang2025video2layout,
  title={Video2Layout: Recall and Reconstruct Metric-Grounded Cognitive Map for Spatial Reasoning},
  author={Huang, Yibin and Xu, Wang and Zhang, Wanyue and Zhi, Helu and Huang, Jingjing and Xu, Yangbin and Sun, Yangang and Zhu, Conghui and Zhao, Tiejun},
  journal={arXiv preprint arXiv:2511.16160},
  year={2025}
}

@inproceedings{yin2025spatial,
  title={Spatial mental modeling from limited views},
  author={Yin, Baiqiao and Wang, Qineng and Zhang, Pingyue and Zhang, Jianshu and Wang, Kangrui and Wang, Zihan and Zhang, Jieyu and Chandrasegaran, Keshigeyan and Liu, Han and Krishna, Ranjay and others},
  booktitle={Structural Priors for Vision Workshop at ICCV'25},
  year={2025}
}

@article{yang2025mindjourney,
  title={MindJourney: Test-Time Scaling with World Models for Spatial Reasoning},
  author={Yang, Yuncong and Liu, Jiageng and Zhang, Zheyuan and Zhou, Siyuan and Tan, Reuben and Yang, Jianwei and Du, Yilun and Gan, Chuang},
  journal={arXiv preprint arXiv:2507.12508},
  year={2025}
}

@article{wang2026hydra,
  title={Hydra-Nav: Object Navigation via Adaptive Dual-Process Reasoning},
  author={Wang, Zixuan and Fang, Huang and Wang, Shaoan and Luo, Yuanfei and Dong, Heng and Li, Wei and Gan, Yiming},
  journal={arXiv preprint arXiv:2602.09972},
  year={2026}
}

\end{document}